\documentclass[conference]{IEEEtran}
\IEEEoverridecommandlockouts
 
\usepackage{cite}
\usepackage{multirow}
\usepackage{amsmath,amssymb,amsfonts}
\usepackage{algorithmic}
\usepackage{graphicx}
\usepackage{textcomp}
\usepackage{xcolor}
\usepackage{hyperref}
\usepackage{graphicx}
\usepackage{booktabs}
\usepackage{multirow}
\usepackage{amsmath}
\usepackage{gensymb}
\usepackage{paralist}
\usepackage{tabularx}
\usepackage{graphicx}
\usepackage{makecell}

\ifCLASSOPTIONcompsoc
\usepackage[caption=false,font=normalsize,labelfont=sf,textfont=sf]{subfig}
\else
\usepackage[caption=false,font=footnotesize]{subfig}
\fi

\hypersetup{
    hidelinks=true
}

\newcommand{\pcrit}{P_\mathcal{R}}
\newcommand{\rcrit}{R_\mathcal{S}}

\def\BibTeX{{\rm B\kern-.05em{\sc i\kern-.025em b}\kern-.08em
    T\kern-.1667em\lower.7ex\hbox{E}\kern-.125emX}}

\begin{document}

\title{Object criticality for safer navigation}

\author{\IEEEauthorblockN{Andrea Ceccarelli}
\IEEEauthorblockA{\textit{Dept. of Mathematics and Informatics} \\
\textit{University of Florence}\\
Florence, Italy \\
andrea.ceccarelli@unifi.it}
\and
\IEEEauthorblockN{Leonardo Montecchi}
\IEEEauthorblockA{\textit{Department of Computer Science} \\
\textit{NTNU}\\
Trondheim, Norway \\
leonardo.montecchi@ntnu.no}
}

\maketitle

\begin{abstract}
Object detection in autonomous driving consists in perceiving and locating instances of objects in multi-dimensional data, such as images or lidar scans. Very recently, multiple works are proposing to evaluate object detectors by measuring their ability to detect the objects that are most likely to interfere with the driving task. Detectors are then ranked according to their ability to detect the most relevant objects, rather than the highest number of objects. However there is little evidence so far that the relevance of predicted object may contribute to the safety and reliability improvement of the driving task. This position paper elaborates on a strategy, together with partial results, to i) configure and deploy object detectors that successfully extract knowledge on object relevance, and ii) use such knowledge to improve the trajectory planning task. We show that, given an object detector, filtering objects based on their relevance, in combination with the traditional confidence threshold, reduces the risk of missing relevant objects, decreases the likelihood of dangerous trajectories, and improves the quality of trajectories in general.
\end{abstract}


\section{Position statement and implications} 
Today, emerging safety-critical technologies rely on object detection as a fundamental part of their perceptual interface to the environments. The most prominent example is automated transportation systems and autonomous driving \cite{premebida2019rgb}:  object detection is a fundamental task for autonomous driving, as it is at the basis of the autonomous driving pipeline. The literature is rather unanimous that the general abstraction of the pipeline scheme of autonomous vehicles first obtains data from the observer and from the object detector, and then, based on such information, a scene representation is created to perform motion  planning \cite{claussmann2019review}, \cite{gonzalez2015review}.  Noteworthy, in the autonomous driving domain, under the name of object detectors are actually comprised perception models that go beyond the mere identification of object location and classification, but that instead also identify additional attributes required for successive  planning tasks, such as object size, distance from the observer, and orientation \cite{grigorescu2020survey}, \cite{teng2023motion}.

In the last four years, context-aware and safety-aware metrics for object detection have been proposed, with the objective to evaluate object detectors with respect to the safety and reliability of the system in which they will operate. Such works have tried to measure criticality in object detection, proposing it as an approach to rank object detectors, in different ways. Lyssenko et al.~\cite{Lyssenko2021} measures the maximum distance at which pedestrians detection does not fail, while Wolf et al.~\cite{Wolf20212759} weights detections according
to the position and estimated time-to-collision with
the object. Volk et al.~\cite{Volk2020} proposes a detection metric that includes 
the relevance of predicted objects with respect to
the observer, Ceccarelli et al.~\cite{Montecchi2022} associate a criticality to each object based on distance from the observer and the estimated velocity, and Topan et al.~\cite{topan2022interaction} present a model to compute safety zones and define safety evaluation metrics for analyzing perception performance of an autonomous vehicle. Further, Bansal et al.~\cite{bansal2021risk} present an improved recall metric for object detection systems, where objects are categorized within three ranks, based on the risk of collision;  Cheng~\cite{cheng2020safety}  formulates the need of ad-hoc detectors to detect objects in  critical areas, where a critical area is the area nearby the observer where failed detection of an object
may lead to immediate safety risks.

Somehow related, other works address the problem of deep neural network uncertainty in autonomous driving, where the term uncertainty should be interpreted in the broad sense of how certain an object detector is about its predicted objects \cite{feng2018towards} and how to learn the uncertainty of a detector \cite{meyer2020learning}. There are numerous variants, as the work of Lo Quercio et al.~\cite{loquercio2020general}, that include information on uncertainty sources (e.g., sensor noise) in the detection. However, the impact of uncertainty to the safety and reliability of the driving tasks is rarely investigated.

The reviewed works rarely or insufficiently demonstrate that their proposed solution actually increases the safety of the planning task, or of the encompassing system, despite it being the driving force that inspired those works. To answer this question, we believe it is necessary to consider trajectory planning, that is a crucial step of the autonomous driving pipeline that comes after object detection. It is evident that the best detection models should make the planner compute a trajectory as close as possible to the one computed using ground truth information \cite{Philion202014052}. A non-optimal trajectory may reduce the quality of the driving task or, in some cases, it may also lead to safety issues. Consequently, it stems from system safety requirements that object detectors should not fail at detecting the elements that are most important for motion planning.

While the impact of object detection to the planning tasks is well understood, only very recently there have been efforts to quantify it with metrics. In particular, the work in \cite{schreier2023offline} and \cite{Philion202014052} evaluates the impact of object detection on the driving performance, with the purpose of proposing metrics to evaluate and compare object detectors. 

\begin{table*}
\caption{Overview of related works and positioning of our contribution}
\label{tab:table1}
\begin{tabularx}{\linewidth}{Xccccccccccccc}
\toprule
&
\cite{Lyssenko2021} &
\cite{Wolf20212759} &
\cite{Volk2020} &
\cite{Montecchi2022} &
\cite{topan2022interaction} &
\cite{bansal2021risk} &
\cite{bansal2022verifiable} &
\cite{feng2018towards} &
\cite{loquercio2020general} &
\cite{cheng2020safety} &
\cite{meyer2020learning} & 
\cite{Philion202014052} &
\textbf{Ours}\\
\hline
{\em Computes uncertainty}
& & & & & & & & $\times$ & $\times$ & & $\times$ & & \\
{\em New metric or new evaluation method}
& $\times$ & $\times$ & $\times$ & $\times$ & $\times$ & $\times$ & $\times$ & & & $\times$ & &$\times$ &\\
{\em Measures and compares object detectors }
& $\times$ & $\times$ & $\times$ & $\times$ & $\times$ & $\times$ & $\times$ & & & & & $\times$ &$\times$ \\
{\em Includes trajectory planning}
& & & & & & & & & $\times$ & & & $\times$ &$\times$ \\
{\em Applicable at runtime}
& & & & & & & & $\times$& & $\times$ & & & $\times$ \\
\bottomrule
\end{tabularx}
\vspace{-2mm}
\normalsize
\end{table*}

Given these premises, we hypothesize that object criticality has a relevance not only for selecting the most suitable object detector, i.e., the most capable in ranking objects, but it is a fundamental support, later in the pipeline, to prioritize objects in situations where detection is uncertain. This is left out in literature, where object criticality is rarely connected to the rest of the pipeline: the benefit of being aware of criticality is not exploited beyond the evaluation of the object detector itself.  As it can be seen in \autoref{tab:table1}, the related works focus merely on the object detection issues, with the limited exceptions of \cite{loquercio2020general}, which shows that the framework can be used to estimate uncertainty on steering prediction without losing prediction performance, and \cite{cheng2020safety}, which proposes to design the object detection module based on the safety requirements of the encompassing system, without however providing experimental results. In \autoref{tab:table1}, we say our approach is applicable at runtime because it filters the predicted objects that are later used in the pipeline.

In other words, in this paper we aim to understand the impact of object criticality on the driving task and its safety, beyond the mere evaluation of object detectors. We hypothesize that object criticality should be connected to the trajectory planning task, so that better (safer) trajectories are computed. More specifically, we propose two main lines of investigation: i) understanding if filtering out less relevant objects can improve trajectory planning; this is based on the assumption that too many irrelevant objects may confuse the trajectory planner; and ii) understanding if retaining objects with high criticality, even if detected with low confidence, can improve safety of the driving tasks; this is based on the assumption that the amount of hazardous false negatives will be reduced. 

We organize our research strategy in three Hypotheses, two of which are partially answered in this paper, while the third one is left for future research.

\textbf{\em Hypothesis 1.} We argue and show that filtering out boxes based on criticality scores, i.e., removing predicted objects that are not relevant for the driving task, has a positive impact on the trajectory planner, meaning that it generally performs better. Our hypothesis is that the reduction of elements avoids creating unnecessary confusion to the planning task. Currently, to the best of our knowledge, this is not known at the state of the art. 

\textbf{\em Hypothesis 2.} Further, we argue that objects detected with low confidence should not be filtered out, if their potential presence may critically affect the driving 
 task. For example, the prediction of a vehicle very close to the observer should not be discarded, regardless of the confidence in the prediction, because it is assigned high criticality. Our hypothesis is that this will reduce the risk of False Negatives of relevant objects, thus improving safety of the driving task, even if at the cost of some additional False Positives. This is an hypothesis that we aim to prove, and to contrast with the by-the-book, widely known, application of confidence thresholds to filter predicted objects. 

\textbf{\em Hypothesis 3.} A consequence of Hypothesis 2 is that there is a relation between confidence threshold and criticality, which may lead to an improvement of the selection of predicted objects that should be fed to trajectory planner. We aim to investigate the benefits of filtering objects based on a combination of confidence threshold and assigned object criticality, where a specific confidence threshold is applied based on the predicted criticality of an object, to decide on the validity of a detection. As Hypothesis 2, this hypothesis is again not confirmed or disproved at the state of the art.


\section{Background}
\subsection{Object Detection}
The task of object detection consists in 
locating and classifying semantic objects of certain object classes within an input
\textit{sample}, with the sample
being either (one or more) 2D visual images or 3D point clouds. The output of an object detection model is a list
of \textit{bounding boxes} (BBs), their \textit{labels} and their \textit{confidence scores} \cite{Jiao2019128837}. The label is the predicted semantic class of the object, and the confidence score reflects the confidence of the detection model in that prediction. BBs are tightly bound boxes encompassing objects in the sample, represented in 2D and 3D as rectangles and cuboids, respectively. Object detectors compute BBs with an assigned \emph{confidence score}. 

\begin{figure}[b!]
  \centering
  \subfloat[All predicted BBs]{\includegraphics[width=.33\linewidth]{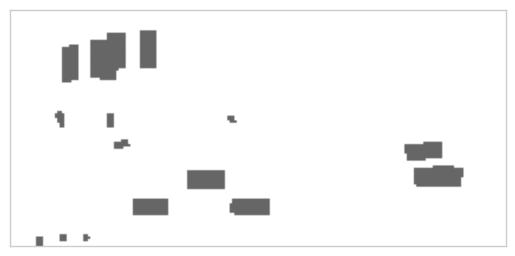}}
  \hfill
  \subfloat[BBs filtered by confidence threshold]{\includegraphics[width=.33\linewidth]{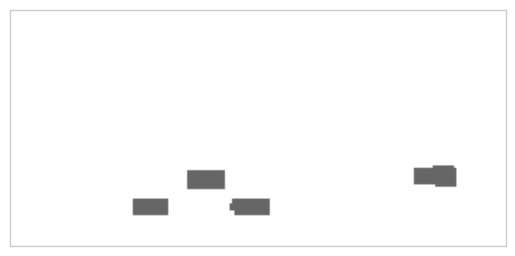}}
  \hfill
  \subfloat[Ground truth BBs]{\includegraphics[width=.33\linewidth]{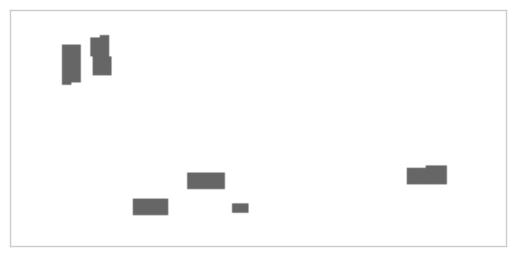}}
\caption{BBs generated by the object detector \cite{SSN} using a sample from \cite{Caesar202011618}. The optimal confidence threshold 0.33 is defined experimentally. 
}
\label{fig:figure1}
\end{figure}

In practice, an object detection produces many bounding boxes, each with a confidence score in the interval $[0,1]$. Then, to retain only the most credible bounding boxes, a \emph{confidence threshold} is applied as a configuration parameter: all BBs with a confidence score above the selected threshold are considered as \emph{predictions}, while the others are discarded.
This is graphically represented in the birdview of \autoref{fig:figure1}.
Noteworthy, models developed for autonomous driving tasks typically do not stop at identifying BBs, but they also determine the kind of object (i.e., they perform classification) and further they compute key attributes like orientation, velocity, and distance from the observer.

\subsection{Object Criticality metrics}
The metrics typically used to evaluate object detectors (precision, recall, and average precision \cite{padilla2020survey}) indicate the ability of ODs to accurately predict instances of objects in a sample, but they do not consider the importance of such objects within the specific scenario. Research on applying ODs in safety-critical environments has raised the problem of defining safety-aware evaluation metrics. In this work, we rely on the solution devised in Ceccarelli et al.~\cite{Montecchi2022}.
The authors 
propose the {Object Criticality Model (OCM)}, in which 
a criticality score is assigned to each object of a specific sample, based on safety-relevant factors relating the object and the observer. Such score is computed for both the ground truth objects and the predicted objects.
To compute the criticality of an object $B$, three factors are considered:
distance, colliding trajectory, and time-to-collision, which result in three individual
criticality scores, $\kappa_d(B)$, $\kappa_r(B)$, and $\kappa_t(B)$, computed for each object.
Each of these scores ranges in the interval $[0,1]$, with 1 meaning maximum
criticality. The model depends on three parameters, $D_{\textrm{max}}$, $R_{\textrm{max}}$, and $T_{\textrm{max}}$, each defining a scaling factor and a threshold after which the corresponding criticality assumes value 0. For example, $D_{\textrm{max}}=30$ means that for objects farther than 30 meters  $\kappa_d(B)=0$. 
The overall criticality weight of an object, $\kappa(B)$, is defined
as a linear combination of the three above weights.

The performance of a detector is then measured in terms of ``how much criticality'' it is able to detect. 
More specifically, OCM includes two metrics called \emph{reliability-weighted precision} ($\pcrit$), and \emph{safety-weighted recall} ($\rcrit$), as variants of the Precision and Recall metrics, with objects weighted based on their criticality score. 

\subsection{Evaluating the quality of trajectory planning}
In \cite{Philion202014052}, Philion et al.~argue that the evaluation of the
performance of perception systems in autonomous vehicles should be aligned with the downstream task of trajectory planning. They propose the Planning KL-divergence (PKL)
metric, as a measure of the difference between the trajectory planned based on
ground truth objects, and the trajectory planned based on objects predicted by an object
detector.

In more details, PKL is a measure of the KL-divergence \cite{kl-divergence} between the probability distribution of future positions of the vehicle, at different time
steps, given the semantic observations (predictions) of the detector and the ideal observations represented by ground truth objects \cite{Philion202014052}. 
Very practically, PKL computes trajectories using all the map data and the predicted objects, and predicts the future position of the observer up to $4$ seconds in the future. As a measure of divergence, a ``perfect'' detector would receive a PKL score of 0, corresponding to no divergence between the trajectories obtained with ground truth objects and with predicted objects. Such divergence is computed for each sample of a target dataset, and then mean and median values of all the samples are computed. 

\section{Approach and implementation}
Our approach relies on the combination of the object criticality scores computed for the BBs using the technique in  \cite{Montecchi2022}, and the PKL metric. Very briefly, this allows us filtering out the object detected in a sample, based on their criticality and the confidence score, and feeding the resulting objects to the PKL function. The output of the PKL function is an indication of the quality of the trajectory computed with the filtered objects. 

However, it should be noted that PKL does not give indication on the safety of the trajectory, i.e., it is only a measure of distance from the ground truth trajectories, but in practice even small differences may have unsafe consequences. Consolidated work will require some more detailed inspection of results beyond the mere computation of PKL, on which we focus on this proposal and that is anticipated below.

\subsection{Experimental setup}
We rely on the nuScenes \cite{Caesar202011618} dataset and on two object detectors from mmdetection3d \cite{mmdet3d2020,chen2019mmdetection}. The first is a recent large-scale dataset for autonomous driving that reports sequences of samples (images and point clouds) collected from a vehicle. Each sequence is 20 seconds long. The latter is an open-source object detection toolbox for 3D detection, that offers state-of-the-art models trained on the nuScenes dataset. We apply the object detectors REG \cite{REGradosavovic2020designing} and FCOS3D \cite{FCOSwang2021fcos3d} on a validation set composed of 10 nuScenes sequences, randomly selected amongst the 150 nuScenes sequences that compose the original validation set. This subset is created due to performance issues: computing pkl under multiple configuration is a time-consuming task. The Average Precision (AP) of the two object detectors on the 10 sequences is respectively $0.45$ and $0.32$. On the whole validation dataset (150 sequences), the AP is very similar, respectively $0.44$ and $0.32$, which means our validation subset can be considered sufficiently representative of the original validation set. 

To compute criticality scores for each predicted object, we manipulate the nuScenes dev-kit \cite{nuScenesDevkit} library. The rest of computation and data analysis is carried out through Jupyter Notebook. Code and instructions to reproduce results are available at \cite{github}.

We note that an  alternative course of action would be to rely on a driving simulator like \cite{dosovitskiy2017carla} to sample realistic scenes, and use a complete autonomous driving pipeline from object detector to the realization of motion actions. However, simulators only provide biased estimates of real-world performance, and, most importantly, such complete pipelines tailored for simulators are exceedingly rare (necessarily excluding end-to-end approaches) and with limited performance.

\subsection{Investigation of Hypothesis 1}

We show that removing predictions that are not relevant for the driving task has a positive impact on the trajectory planner, i.e., it improves driving quality. We measure the optimal confidence threshold: we experimentally test multiple confidence thresholds, until we find the one that provides the best (the lowest) PKL. This leads to the results in the upper part of \autoref{tab:table2}.
This is a required step for the usage of PKL, as well as a common approach to the configuration of any object detector. 

Next, we filter the predicted objects with confidence score above the confidence threshold, i.e., we start from the predicted objects that are naturally used to compute PKL. We discard the predicted object with a criticality score below a threshold (as for the confidence threshold, the optimal criticality threshold is defined experimentally). Finally, we compute PKL using the remaining predicted objects. We show results in the lower part of \autoref{tab:table2}. 

\renewcommand{\tabcolsep}{4pt}

\begin{table}
\caption{Median and mean PKL computed on 10 nuScenes sequences, (top) selecting objects based only on the optimal confidence threshold, and (bottom) using also the criticality threshold.}
\label{tab:table2}

\begin{tabular}{lcc|cc}
\toprule
& \makecell{confidence  \\ threshold} & \makecell{median \\ PKL} &  \makecell{confidence \\ threshold }& \makecell{ mean \\ PKL} \\
\hline
REG    & 0.55                 & 22.749       & 0.55                 & 78.951     \\
FCOS3D & 0.25                 & 0.986        & 0.15                 & 4.104      \\
\\
       & \makecell{confidence and\\ criticality thresholds} & \makecell{median \\ PKL} & \makecell{confidence and \\  criticality thresholds}  & \makecell{mean \\ PKL} \\
\hline
REG    & [0.55; 0.65]                  & 9.044        & [0.55; 0.60]                  & 43.801     \\
FCOS3D & [0.25; 0.30]                  & 0.966        & [0.15; 0.15]                  & 4.080      \\
\bottomrule
\end{tabular}
\end{table}

The objects that are discarded because of low criticality are either not affecting the driving tasks, or their removal has a positive impact on the trajectory planner. This last case is especially evident with the REG object detector. Note that PKL is logarithmic, and a relatively small reduction of PKL when it is close to zero is actually a relevant improvement. It should be noted that, despite the average precision of FCOS3D is lower (worse) than the one of REG, the PKL of FCOS3D is significantly better than the one measured with REG. This should not confuse the reader: average precision is based on the ability to detect BBs and classify the object type, while PKL relies on  additional semantic, namely the size, detected orientation and velocity of the predicted objects. This observation is also an evidence  that measuring average precision, precision and recall is not indicative on the quality of the object detector to support the autonomous driving pipeline.

\subsection{Investigation of Hypothesis 2}
 
We avoid removing objects with a high predicted criticality, independently on the confidence score assigned by the object detector. Technically, the implementation is as follows: i) first, all the predicted objects with predicted criticality above a defined threshold are kept; ii) the remaining objects are kept only if the confidence score is above the confidence threshold.

Our experimental results show that PKL is not improving. For example, the best mean and median PKL we obtain with REG are respectively 81.364 and 27.325, with a criticality threshold of 0.95 (which identifies only the objects that are in a close proximity to the observer). This result is reasonable, considering that we maintain \emph{every} predicted object that is potentially dangerous for the navigation of the observer. This includes many predictions of non-existing objects, with the result that the trajectory is modified unnecessarily. Summarizing, the resulting driving experience would be inefficient and unpleasant, as witnessed by the higher PKL.

\begin{figure}
  \centering
  \subfloat[]{\includegraphics[trim={0 0 0 2cm},clip, width=.33\linewidth]{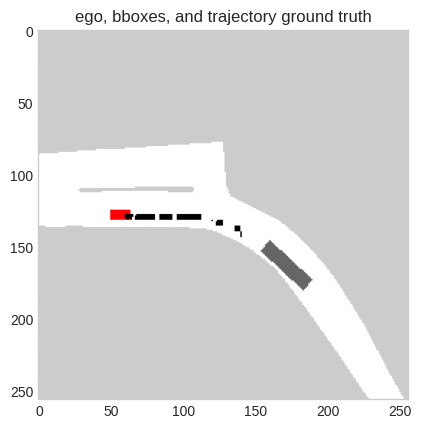}}
  \hfill
  \subfloat[]{\includegraphics[trim={0 0 0 2cm},clip,width=.33\linewidth]{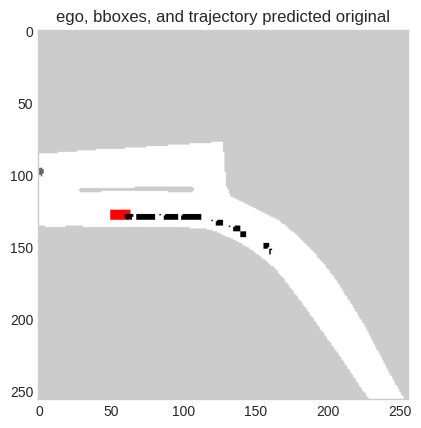}}
  \hfill
  \subfloat[]{\includegraphics[trim={0 0 0 2cm},clip,width=.33\linewidth]{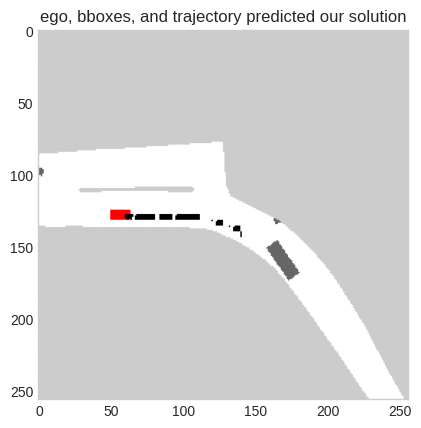}}
\caption{a) ground truth BBs; b) BBs predicted by REG; c) we avoid removing BBs with criticality score above 0.8.}
\label{fig:figure2}
\end{figure}

However, selected examples can show that safety of the navigation is improved. We discuss this with \autoref{fig:figure2}, where we show the birdview of a nuScenes sample. \autoref{fig:figure2}a shows the ground truth, with the observer in red, the trajectory in black, and the existing object (specifically, a bus) in dark grey. In \autoref{fig:figure2}b, we observe that the object detector REG, configured with its optimal confidence threshold 0.55, does not detect the bus. Actually, the bus is amongst the prediction of REG, but it has a confidence score lower than 0.55. In \autoref{fig:figure2}c we plot the BBs according to the criterion explored in Hypothesis 2: all the objects with high criticality score (above 0.8 in this example) are retained, independently on their  confidence score. In this case the bus is included among the predicted objects, because of its potential criticality in the sample. The bus is detected with a smaller size and slightly wrong orientation, but the detection is sufficiently precise to adjust the trajectory: as we can see, the trajectory of \autoref{fig:figure2}c is closer to the ground truth than the trajectory of \autoref{fig:figure2}b, and it is intuitively safer as it is computed being aware of the existence of the bus.

\subsection{Investigation of Hypothesis 3}
From the discussion above, it is reasonable to infer that maintaining  all predicted objects above a criticality threshold is safe but inefficient. We aim to relax the condition that all the objects above a criticality threshold are maintained. 

Practically, this can be realized by setting multiple pairs of confidence threshold and criticality threshold, or with more complicated approaches, e.g., by identifying functions that relate the criticality and confidence scores. However, at the present stage of our research, we do not have a preferred approach to propose and related experiments.


It is also necessary to discuss that our ultimate goal is improving safety, without significantly reducing the reliability of the driving task. While the latter can somehow be measured using PKL, the safety of the driving task needs to be measured in a different way, as the counterexample of \autoref{fig:figure2} showed. 

Given a sample at any time instant $t$, we propose to compute the trajectory up to $t+4$ seconds in the future, as done by PKL, and project it on a birdview which displays the observer at time $t$, the trajectory computed, and the ground truth BBs of all vehicles within a window of $[t; t+4]$ seconds. If, within any moment in that time window, there are overlaps between the trajectory and the ground truth BBs, we can suspect that there is a violation of safety (i.e., a possible collision), and we can also visually inspect the sample for confirmation. 

Very practically, this can be realized by modifying the functionality described in \cite{10.1007/978-3-030-65414-6_2}. We are able to extract a birdview map of the planned trajectory, and the ground truth position of each vehicle, after $t$ time, with $t \in [0,4]$ seconds (with 16 intervals of 0.25 seconds). In this way, if there are overlaps, we can claim a safety hazard. There are some technical limitations to this approach that need to be solved. In fact, PKL does not provide a unique trajectory, but a map of positions; thus, we can only extract the most probable trajectory as the set of positions with the highest probability. This approach is also used in \cite{10.1007/978-3-030-65414-6_2} to draw trajectories. Second, an overlap between ground truths and trajectory within $4$ seconds does not imply that in practice a dangerous situation will occur. It is in fact challenging to provide a proper measure of safety in this context, and this will be part of our future work. Besides numerical evaluation, we aim to validate the hypotheses stated in this paper by involving humans: individuals with driving experience and safety experts. Ideally, the trajectories based on objects filtered with our method should be considered safer also from the user's perspective.





\section*{Acknowledgments}
This work has been partially supported by project SERICS (PE00000014) under the MUR National Recovery and Resilience Plan funded by the European Union - NextGenerationEU, and by the PRIN 2022 project FLEGREA (B53D23012930006) and the PRIN PNRR 2022 project BREADCRUMBS (P2022K7ERB) funded by the Italian Ministry of University and Research.

\bibliographystyle{IEEEtran}
\bibliography{bibliography}

\begin{thebibliography}{10}
\providecommand{\url}[1]{#1}
\csname url@samestyle\endcsname
\providecommand{\newblock}{\relax}
\providecommand{\bibinfo}[2]{#2}
\providecommand{\BIBentrySTDinterwordspacing}{\spaceskip=0pt\relax}
\providecommand{\BIBentryALTinterwordstretchfactor}{4}
\providecommand{\BIBentryALTinterwordspacing}{\spaceskip=\fontdimen2\font plus
\BIBentryALTinterwordstretchfactor\fontdimen3\font minus \fontdimen4\font\relax}
\providecommand{\BIBforeignlanguage}[2]{{%
\expandafter\ifx\csname l@#1\endcsname\relax
\typeout{** WARNING: IEEEtran.bst: No hyphenation pattern has been}%
\typeout{** loaded for the language `#1'. Using the pattern for}%
\typeout{** the default language instead.}%
\else
\language=\csname l@#1\endcsname
\fi
#2}}
\providecommand{\BIBdecl}{\relax}
\BIBdecl

\bibitem{premebida2019rgb}
C.~Premebida, G.~Melotti, and A.~Asvadi, ``Rgb-d object classification for autonomous driving perception,'' \emph{RGB-D Image Analysis and Processing}, pp. 377--395, 2019.

\bibitem{claussmann2019review}
L.~Claussmann, M.~Revilloud, D.~Gruyer, and S.~Glaser, ``A review of motion planning for highway autonomous driving,'' \emph{IEEE Transactions on Intelligent Transportation Systems}, vol.~21, no.~5, pp. 1826--1848, 2019.

\bibitem{gonzalez2015review}
D.~Gonz{\'a}lez, J.~P{\'e}rez, V.~Milan{\'e}s, and F.~Nashashibi, ``A review of motion planning techniques for automated vehicles,'' \emph{IEEE Transactions on intelligent transportation systems}, vol.~17, no.~4, pp. 1135--1145, 2015.

\bibitem{grigorescu2020survey}
S.~Grigorescu, B.~Trasnea, T.~Cocias, and G.~Macesanu, ``A survey of deep learning techniques for autonomous driving,'' \emph{Journal of Field Robotics}, vol.~37, no.~3, pp. 362--386, 2020.

\bibitem{teng2023motion}
S.~Teng, X.~Hu, P.~Deng, B.~Li, Y.~Li, Y.~Ai, D.~Yang, L.~Li, Z.~Xuanyuan, F.~Zhu \emph{et~al.}, ``Motion planning for autonomous driving: The state of the art and future perspectives,'' \emph{IEEE Transactions on Intelligent Vehicles}, 2023.

\bibitem{Lyssenko2021}
M.~Lyssenko, C.~Gladisch, C.~Heinzemann, M.~Woehrle, and R.~Triebel, ``From evaluation to verification: Towards task-oriented relevance metrics for pedestrian detection in safety-critical domains,'' in \emph{2021 IEEE/CVF Conference on Computer Vision and Pattern Recognition Workshops (CVPRW)}, 2021, pp. 38--45.

\bibitem{Wolf20212759}
M.~Wolf, L.~R. Douat, and M.~Erz, ``Safety-aware metric for people detection,'' in \emph{IEEE Conference on Intelligent Transportation Systems, Proceedings, ITSC}, vol. 2021-September, 2021, pp. 2759--2765.

\bibitem{Volk2020}
G.~Volk, J.~Gamerdinger, A.~V. Betnuth, and O.~Bringmann, ``A comprehensive safety metric to evaluate perception in autonomous systems,'' in \emph{2020 IEEE 23rd International Conference on Intelligent Transportation Systems, ITSC 2020}, 2020.

\bibitem{Montecchi2022}
A.~Ceccarelli and L.~Montecchi, ``{Evaluating Object (Mis)Detection From a Safety and Reliability Perspective: Discussion and Measures},'' \emph{IEEE Access}, vol.~11, pp. 44\,952--44\,963, 5 2023.

\bibitem{topan2022interaction}
S.~Topan, K.~Leung, Y.~Chen, P.~Tupekar, E.~Schmerling, J.~Nilsson, M.~Cox, and M.~Pavone, ``Interaction-dynamics-aware perception zones for obstacle detection safety evaluation,'' in \emph{2022 IEEE Intelligent Vehicles Symposium (IV)}.\hskip 1em plus 0.5em minus 0.4em\relax IEEE, 2022, pp. 1201--1210.

\bibitem{bansal2021risk}
A.~Bansal, J.~Singh, M.~Verucchi, M.~Caccamo, and L.~Sha, ``Risk ranked recall: Collision safety metric for object detection systems in autonomous vehicles,'' in \emph{2021 10th Mediterranean Conference on Embedded Computing (MECO)}.\hskip 1em plus 0.5em minus 0.4em\relax IEEE, 2021, pp. 1--4.

\bibitem{cheng2020safety}
C.-H. Cheng, ``Safety-aware hardening of 3d object detection neural network systems,'' in \emph{Computer Safety, Reliability, and Security: 39th International Conference, SAFECOMP 2020, Lisbon, Portugal, September 16--18, 2020, Proceedings 39}.\hskip 1em plus 0.5em minus 0.4em\relax Springer, 2020, pp. 213--227.

\bibitem{feng2018towards}
D.~Feng, L.~Rosenbaum, and K.~Dietmayer, ``Towards safe autonomous driving: Capture uncertainty in the deep neural network for lidar 3d vehicle detection,'' in \emph{2018 21st international conference on intelligent transportation systems (ITSC)}.\hskip 1em plus 0.5em minus 0.4em\relax IEEE, 2018, pp. 3266--3273.

\bibitem{meyer2020learning}
G.~P. Meyer and N.~Thakurdesai, ``Learning an uncertainty-aware object detector for autonomous driving,'' in \emph{2020 IEEE/RSJ International Conference on Intelligent Robots and Systems (IROS)}.\hskip 1em plus 0.5em minus 0.4em\relax IEEE, 2020, pp. 10\,521--10\,527.

\bibitem{loquercio2020general}
A.~Loquercio, M.~Segu, and D.~Scaramuzza, ``A general framework for uncertainty estimation in deep learning,'' \emph{IEEE Robotics and Automation Letters}, vol.~5, no.~2, pp. 3153--3160, 2020.

\bibitem{Philion202014052}
J.~Philion, A.~Kar, and S.~Fidler, ``Learning to evaluate perception models using planner-centric metrics,'' in \emph{Proceedings of the IEEE Computer Society Conference on Computer Vision and Pattern Recognition}, 2020, pp. 14\,052--14\,061.

\bibitem{schreier2023offline}
T.~Schreier, K.~Renz, A.~Geiger, and K.~Chitta, ``On offline evaluation of 3d object detection for autonomous driving,'' in \emph{Proceedings of the IEEE/CVF International Conference on Computer Vision}, 2023, pp. 4084--4089.

\bibitem{bansal2022verifiable}
A.~Bansal, H.~Kim, S.~Yu, B.~Li, N.~Hovakimyan, M.~Caccamo, and L.~Sha, ``Verifiable obstacle detection,'' in \emph{2022 IEEE 33rd International Symposium on Software Reliability Engineering (ISSRE)}.\hskip 1em plus 0.5em minus 0.4em\relax IEEE, 2022, pp. 61--72.

\bibitem{Jiao2019128837}
L.~Jiao, F.~Zhang, F.~Liu, S.~Yang, L.~Li, Z.~Feng, and R.~Qu, ``A survey of deep learning-based object detection,'' \emph{IEEE Access}, vol.~7, pp. 128\,837--128\,868, 2019.

\bibitem{SSN}
X.~Zhu, Y.~Ma, T.~Wang, Y.~Xu, J.~Shi, and D.~Lin, ``{SSN: Shape Signature Networks for Multi-class Object Detection from Point Clouds},'' in \emph{Computer Vision – ECCV 2020}, vol. 12370 LNCS, 2020, pp. 581--597.

\bibitem{Caesar202011618}
H.~Caesar, V.~Bankiti, A.~H. Lang, S.~Vora, V.~E. Liong, Q.~Xu, A.~Krishnan, Y.~Pan, G.~Baldan, and O.~Beijbom, ``Nuscenes: A multimodal dataset for autonomous driving,'' in \emph{Proceedings of the IEEE Computer Society Conference on Computer Vision and Pattern Recognition}, 2020, pp. 11\,618--11\,628.

\bibitem{padilla2020survey}
R.~Padilla, S.~L. Netto, and E.~A. Da~Silva, ``A survey on performance metrics for object-detection algorithms,'' in \emph{2020 international conference on systems, signals and image processing (IWSSIP)}.\hskip 1em plus 0.5em minus 0.4em\relax IEEE, 2020, pp. 237--242.

\bibitem{kl-divergence}
T.~Kim, J.~Oh, N.~Kim, S.~Cho, and S.-Y. Yun, ``{Comparing Kullback-Leibler Divergence and Mean Squared Error Loss in Knowledge Distillation},'' arXiv 2105.08919, 2021.

\bibitem{mmdet3d2020}
{MMDetection3D Contributors}, ``{MMDetection3D: OpenMMLab} next-generation platform for general {3D} object detection,'' \url{https://github.com/open-mmlab/mmdetection3d} (Accessed: June 7, 2023).

\bibitem{chen2019mmdetection}
K.~Chen, J.~Wang, J.~Pang, Y.~Cao, Y.~Xiong, X.~Li, S.~Sun, W.~Feng, Z.~Liu, J.~Xu, Z.~Zhang, D.~Cheng, C.~Zhu, T.~Cheng, Q.~Zhao, B.~Li, X.~Lu, R.~Zhu, Y.~Wu, J.~Dai, J.~Wang, J.~Shi, W.~Ouyang, C.~C. Loy, and D.~Lin, ``{MMDetection: Open MMLab Detection Toolbox and Benchmark},'' arXiv 1906.07155, 2019.

\bibitem{REGradosavovic2020designing}
I.~Radosavovic, R.~P. Kosaraju, R.~Girshick, K.~He, and P.~Dollár, ``Designing network design spaces,'' 2020.

\bibitem{FCOSwang2021fcos3d}
T.~Wang, X.~Zhu, J.~Pang, and D.~Lin, ``Fcos3d: Fully convolutional one-stage monocular 3d object detection,'' in \emph{Proceedings of the IEEE/CVF International Conference on Computer Vision (ICCV) Workshops}, October 2021, pp. 913--922.

\bibitem{nuScenesDevkit}
{nuScenes Contributors}, ``{nuScenes devkit},'' \url{https://github.com/nutonomy/nuscenes-devkit} (Accessed: June 7, 2023).

\bibitem{github}
\emph{A. Ceccarelli, L. Montecchi}, ``Github repository \url{https://github.com/AndreaCecca/detectorAndTrajectory},'' online, 2023.

\bibitem{dosovitskiy2017carla}
A.~Dosovitskiy, G.~Ros, F.~Codevilla, A.~Lopez, and V.~Koltun, ``Carla: An open urban driving simulator,'' in \emph{Conference on robot learning}.\hskip 1em plus 0.5em minus 0.4em\relax PMLR, 2017, pp. 1--16.

\bibitem{10.1007/978-3-030-65414-6_2}
J.~Philion, A.~Kar, and S.~Fidler, ``Implementing planning kl-divergence,'' in \emph{Computer Vision -- ECCV 2020 Workshops}, A.~Bartoli and A.~Fusiello, Eds.\hskip 1em plus 0.5em minus 0.4em\relax Cham: Springer International Publishing, 2020, pp. 11--18.

\end{thebibliography}
\end{document}